\documentclass[lettersize,journal]{IEEEtran}
\usepackage{amsmath,amsfonts}
\usepackage{array}
\usepackage[caption=false,font=normalsize,labelfont=sf,textfont=sf]{subfig}
\usepackage{textcomp}
\usepackage{stfloats}
\usepackage{url}
\usepackage{verbatim}
\usepackage{graphicx}
\usepackage{cite}
\usepackage{times}
\usepackage{soul}
\usepackage{url}
\usepackage[hidelinks]{hyperref}
\usepackage[utf8]{inputenc}
\usepackage[small]{caption}
\usepackage{graphicx}
\usepackage{amsmath}
\usepackage{amsthm}
\usepackage{booktabs}
\usepackage{algorithm}
\usepackage{algpseudocode}

\usepackage[switch]{lineno}

\usepackage{amsmath}
\usepackage{amssymb}
\usepackage{array}
\usepackage{graphicx}
\usepackage{graphicx}
\usepackage{float}
\usepackage{longtable}
\usepackage{svg}
\usepackage{amsthm}
\usepackage{subfig}

\usepackage{multirow}

\newtheorem{theorem}{Theorem}
\newtheorem{assumption}{Assumption}
\newtheorem{corollary}{Corollary}

\usepackage{caption}
\begin{document}

\title{DPFNAS: Differential Privacy-Enhanced Federated Neural Architecture Search for 6G Edge Intelligence}

\author{Yang Lv, Jin Cao, Ben Niu, Zhe Sun, Fengwei Wang, Fenghua Li, Hui Li

\thanks{}
\thanks{}}

\author{Yang Lv, Jin Cao$^{\ast}$, Ben Niu, Zhe Sun, Fengwei Wang, Fenghua Li, Hui Li%
\thanks{Yang Lv is with the School of Cyber Engineering, Xidian University, Xi'an, China (e-mail: lyuyang@stu.xidian.edu.cn). Jin Cao$^{\ast}$, Fengwei Wang, and Hui Li are with the same affiliation (e-mail: \{jcao, wangfengwei, lihui\}@xidian.edu.cn).}
\thanks{Ben Niu and Fenghua Li are with the Institute of Information Engineering, Chinese Academy of Sciences, Beijing, China (e-mail: \{niuben, lifenghua\}@iie.ac.cn).}%
\thanks{Zhe Sun is with the Cyberspace Institute of Advanced Technology, Guangzhou University, Guangzhou, China (e-mail: sunzhe@gzhu.edu.cn).}%
\thanks{$^{\ast}$Corresponding author: Jin Cao.}%
}



\maketitle

\begin{abstract}

The Sixth-Generation (6G) network envisions pervasive artificial intelligence (AI) as a core goal, enabled by edge intelligence through on-device data utilization. To realize this vision, federated learning (FL) has emerged as a key paradigm for collaborative training across edge devices. 
However, the sensitivity and heterogeneity of edge data pose key challenges to FL: parameter sharing risks data reconstruction, and a unified global model struggles to adapt to diverse local distributions.
In this paper, we propose a novel federated learning framework that integrates personalized differential privacy (DP) and adaptive model design. To protect training data, we leverage sample-level representations for knowledge sharing and apply a personalized DP strategy to resist reconstruction attacks. To ensure distribution-aware adaptation under privacy constraints, we develop a privacy-aware neural architecture search (NAS) algorithm that generates locally customized architectures and hyperparameters.
To the best of our knowledge, this is the first personalized DP solution tailored for representation-based FL with theoretical convergence guarantees.
Our scheme achieves strong privacy guarantees for training data while significantly outperforming state-of-the-art methods in model performance. Experiments on benchmark datasets such as CIFAR-10 and CIFAR-100 demonstrate that our scheme improves accuracy by 6.82\% over the federated NAS method PerFedRLNAS, while reducing model size to 1/10 and communication cost to 1/20.

\end{abstract}


\section{Introduction}

\IEEEPARstart{A}{ccording} to the International Telecommunication Union (ITU), the Sixth-Generation (6G) mobile communication network are expected to fundamentally reshape current network architectures \cite{ITU_R_M2160_2023}. This transformation will be driven by an unprecedented degree of connectivity. 6G networks are expected to support tens of millions of edge device connections per square kilometer  \cite{duan2023combining}. These edge devices—such as smartphones, wearables, and sensors—will continuously generate vast volumes of local data. These data, rich in contextual information and latent intelligence, are key enablers for delivering efficient and responsive artificial intelligent (AI) services.

Nowadays, the utilization of data generated at the edge is still significantly limited in the Fifth-Generation mobile communication system (5GS). In 5GS, the use of AI remains largely decoupled from the network's native architecture \cite{3GPP_TR33_784_2024}. Most current 5G systems rely on add-on modules in the core network, such as the Network Data Analytics Function (NWDAF), to support AI-based functionalities \cite{3GPP_TS23_288_R16}. These modules mainly target network operation tasks, including Quality of Service (QoE) optimization, mobility prediction, and fault diagnosis, using aggregated data from the core network \cite{3GPP_TR33_866_R17, 3GPP_TR33_867_R17, 3GPP_TR28_809_R17}. However, when edge devices require intelligent services, their locally generated data must typically be uploaded to the cloud for centralized processing. This cloud-centric paradigm leads to high communication overhead and introduces latency \cite{shi2020communication}, making it difficult to support real-time, low-latency, and personalized service requirements.

In response, the 6G mobile communication system introduces the core vision of pervasive AI \cite{ITU_R_M2160_2023}, shifting away from the cloud-centric paradigm of 5G toward a fundamentally new architecture based on edge intelligence. In this paradigm, AI is embedded as a native capability within the network, enabling edge devices not only to sense and transmit data, but also to actively participate in model training and inference. This allows large-scale collaboration among edge devices, forming a new generation of distributed, self-adaptive, and intelligence-driven networks \cite{yang2022federated}. Within this architecture, Federated Learning (FL) \cite{mcmahan2017communication} is recognized by the 3rd Generation Partnership Project (3GPP) as a key collaborative training paradigm for realizing edge intelligence in 6G networks, and has already been specified in the 3GPP standard documents \cite{3GPP_TR33_898_R18}.

\begin{figure}
    \centering
    \includegraphics[width=1\linewidth]{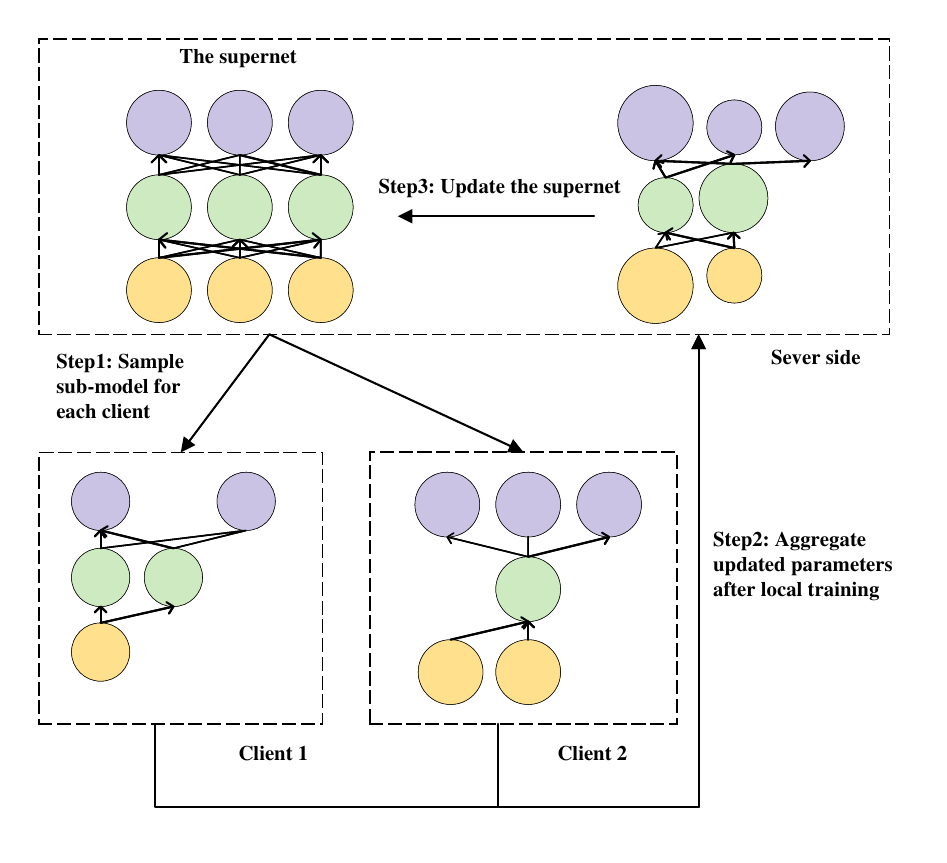}
    \caption{The general workflow of supernet strategy}
    \label{fig:Supernet}
\end{figure}

Although FL is particularly suited to the resource-constrained nature of edge devices, the heterogeneity of data distributions across devices in 6G networks poses significant challenges. A global, unified model architecture often fails to adapt effectively to such variability, which can result in poor convergence and limited generalization performance \cite{zhao2018federated}.
Therefore, how to provide data-specific architectures for clients within FL frameworks has become an urgent problem. 
Recently, Neural Architecture Search (NAS) \cite{elsken2019neural} has emerged as a promising solution for designing data-specific architectures in FL. This type of FL framework design is commonly referred to as Federated Neural Architecture Search (FNAS) \cite{zhu2021federated}. The supernet is a widely adopted strategy to realize NAS in FNAS frameworks design \cite{mushtaq2021spider, zhu2021real, hoang2021personalized, yao2024perfedrlnas, yan2024peaches, wang2023automated}. The general workflow of the supernet strategy is shown in Fig. \ref{fig:Supernet}. The supernet serves as a unified architectural container, defining a shared and controllable search space. Clients are allowed to sample sub-models from this supernet to suit their local data distributions, and these sub-model updated parameters can be aggregated on the server side via structural alignment or masked averaging. By predefining the range of architecture choices for sub-models, the supernet strategy supports the aggregation of heterogeneous sub-model updated parameters, demonstrating compatibility with the FL paradigm that sharing only model updates. 

Nevertheless, supernet-based FNAS approaches face the following two inherent limitations that restrict their applicability in 6G environments: privacy concerns arising from the sharing of updated parameters, and training inefficiency caused by limited model adaptability. Firstly, FL relies on sharing model updates to enable collaborative training \cite{mcmahan2017communication}. In supernet-based FNAS, the shared model updates are the updated parameters \cite{cha2022survey}. These updated parameters are particularly susceptible to being exploited by attackers to infer private information from the data \cite{lyu2020threats}, which contradicts the strict privacy requirements of edge devices in 6G networks. Secondly, to support the aggregation of heterogeneous sub-models, supernet-based FNAS predefines the range of architecture choices for sub-models. However, due to the highly heterogeneous nature of local data across edge devices in 6G networks, some devices may fail to obtain optimal architectures tailored to their local data distributions, resulting in low accuracy and slow convergence \cite{yan2024peaches}. 

To address the above two challenges,  we propose a novel FL framework named Differential Privacy-Enhanced FNAS (DPFNAS). Our objective is to develop a FL framework that simultaneously ensures privacy preservation and supports flexible model adaptability.
Our main contributions can be summarized as follows:

\begin{itemize}
    \item We propose DPRBFT, a personalized, privacy-preserving collaborative training strategy for heterogeneous models, which supports personalized privacy guarantees to further mitigate the inherent privacy risks of representation-based methods. To the best of our knowledge, this is the first work to apply personalized differential privacy to protect intermediate representations in representation-based federated training paradigms. In addition, the use of sample-level intermediate representations improves training robustness under system heterogeneity compared to existing representation-based approaches.

    \item We propose PANAS, a privacy-aware and flexible NAS algorithm that enables optimal architectures tailored to clients' data distributions. By designing a novel lightweight block-based search space, the proposed PANAS enhances the flexibility of architecture design and ensures lightweight models suitable for edge environments. Moreover, the proposed PANAS further leverages Bayesian Optimization to perform automated hyperparameter tuning under differential privacy constraints. This design eliminates the need for additional architecture-specific usability enhancements and can be directly integrated into existing differentially private machine learning methods.

    \item We provide a theoretical convergence analysis of our differentially private training process, ensuring the soundness and stability of the proposed framework. Extensive experiments on widely used classification datasets demonstrate that the proposed DPFNAS outperforms state-of-the-art methods in terms of accuracy, convergence speed, communication efficiency, and privacy protection.
\end{itemize}

The remainder of this paper is organized as follows. Section II reviews related work. Section III details the proposed DPFNAS framework. Section IV presents the experimental setup and results. Finally, Section V concludes the paper and outlines directions for future research.

\section{Related Work}

In this section, we briefly review the research on NAS for client-adaptive modeling in FL, representation-based personalization strategies that decouple global aggregation from local feature extraction, and differential privacy mechanisms for protecting sensitive information during training. We focus on their core methodologies and limitations in supporting heterogeneous models, enabling architectural flexibility, and ensuring privacy-preserving adaptability.

\subsection{NAS in FL}
To overcome the limitations of fixed architectures in FL, the NAS has been introduced to automatically discover performant architectures tailored to local data distributions. Several studies have explored this integration, mostly based on the supernet paradigm, where a shared search space or model pool is leveraged across clients.

FedNAS~\cite{li2020fednas} is one of the early attempts in this direction, where each client searches for optimal weights and architecture parameters within a shared global search space. These parameters are aggregated on the server and then broadcasted back to all clients for the next training round. However, this scheme does not support heterogeneous clients with diverse data and computational capabilities.

To address personalization, SPIDER~\cite{mushtaq2021spider} maintains both a global supernet and a local sub-network for each client. Clients alternate between global model training and local adaptation. However, the need for each client to store and train on the full supernet incurs significant resource overhead.

RT-FedEvoNAS~\cite{zhu2021real} introduces a sub-network sampling strategy, where each client trains only a randomly selected sub-model from a global supernet. This design significantly reduces communication costs by limiting the size of transmitted models. Nevertheless, such random sampling can lead to sub-models misaligned with local data, and weight inheritance may exacerbate the mismatch.

Peaches~\cite{yan2024peaches} extends the supernet-based paradigm by allowing each client to maintain a hybrid supernet composed of a globally shared base and a client-specific extension. During training, only the shared base is synchronized across clients, enabling partial personalization while reducing communication overhead. However, the design of this shared component remains tightly limited by the communication bandwidth and hardware constraints of the least capable client, thereby restricting overall model expressiveness.

To improve architecture selection, PerFedRLNAS~\cite{yao2024perfedrlnas} employs a reinforcement learning-based controller to select suitable sub-models per round based on reward signals. While this method enhances personalization, the strategy limits search efficiency by only dispatching a single sub-model to each client per communication round.

Despite their progress, existing FL-NAS approaches face intrinsic limitations due to the supernet-based design. Firstly, the reliance on full supernets leads to high communication and storage costs, which are unsuitable for resource-constrained environments. Secondly, the central server typically has access to all model architectures and parameters, raising privacy concerns. Finally, the sub-network adaptability is restricted by the pre-defined structure of the supernet, limiting its flexibility in heterogeneous FL settings.

\subsection{Representation-based FL}

To address the challenges of statistical heterogeneity and model diversity in FL, representation-based FL frameworks have emerged as an alternative to fully shared model training. These methods decouple global collaboration from personalized representation learning, offering a promising paradigm for efficient and adaptive personalization.

FedPer~\cite{arivazhagan2019fedper} pioneers the base-plus-personalization structure in FL, splitting models into shared base layers and client-specific personalization layers. Although this method is effective under statistical heterogeneity, it requires all clients to share the same base architecture, limiting its applicability in heterogeneous settings.

To alleviate this constraint, FedProto~\cite{tan2022fedproto} proposes a prototype-sharing strategy that aggregates class-wise mean feature vectors across clients. This method is inherently agnostic to model architecture, enabling collaboration among fully heterogeneous clients. However, this prototype averaging strategy  may overlook intra-class variance and lacks alignment at the decision boundary level.

In contrast, FedClassAvg~\cite{jang2023fedclassavg} introduces a classifier averaging strategy, wherein only the top-layer parameters are exchanged and aggregated, allowing for locally heterogeneous feature extractors. It further employs supervised contrastive learning to enhance local feature discrimination. Nevertheless, it relies on a unified classifier head and consistent label space, which reduces its flexibility under label-heterogeneous scenarios.

FedGH~\cite{yi2023fedgh} extends this idea by training a global prediction head shared across clients with heterogeneous extractors, based on class-wise averaged representations. This scheme improves generalization and reduces communication cost, but it assumes semantic consistency and thus lacks downstream personalization.

To reduce the communication overhead, existing representation-based FL methods commonly adopt prototype representations, where a class-wise prototype is computed as the average of local representations within the same category. This approach is effective when data are non-IID in a class-wise manner and helps reduce communication costs. However, when the data distribution becomes more dispersed or fine-grained, such averaging can lead to significant information loss.

\subsection{Differential Privacy in FL}

Differential privacy has become a cornerstone of secure FL, but its integration introduces fundamental trade-offs between convergence, utility, and calibration. In the context of differentially private deep learning, the privacy budget $\epsilon$ refers to the maximum allowable cumulative privacy leakage throughout training. Each gradient update incurs a privacy loss, a random variable that quantifies the privacy leakage per step. The privacy cost is defined as the cumulative upper bound of the privacy loss over all training steps. As long as the privacy cost remains below the privacy budget, the training procedure is considered to satisfy the differential privacy guarantee.

NbAFL~\cite{wei2020nbafl} proposes a noise-before-aggregation scheme and provides theoretical convergence bounds, highlighting the privacy-utility trade-off and client selection strategies. 
Personalized DP-FL~\cite{hu2020personalized} addresses personalization under device heterogeneity by integrating multitask learning with differential privacy guarantees, and applies the moment accountant technique—a method for tracking cumulative privacy loss over multiple training iterations—to analyze and manage the privacy budget.

To support fine-grained privacy protection, HPFL~\cite{wu2021hpfl} proposes a hierarchical aggregation mechanism that separates public and private components of user data, offering differentiated privacy protection. In the domain of split learning, MaskSL~\cite{wang2023masksl} incorporates DP into large-scale masked vision pretraining with client-side Fisher information masking.

Beyond algorithm design, the scheme in ~\cite{bu2023calibration} analyzes the convergence and calibration behavior of DP deep learning. It shows that gradient clipping is the dominant factor affecting performance, and large clipping norms improve calibration without reducing accuracy. Zhang et al.~\cite{zhang2021clipping} further provide convergence analysis of client-level DP in FL, linking clipping bias to the distribution of local updates.

Nevertheless, existing works lack convergence theory and utility modeling under the representation-based paradigm, limiting their applicability in real-world FL systems with model partitioning and privacy constraints.

\section{PROPOSED SCHEME: DPFNAS}

In this section, we detail the proposed scheme. By our scheme, each client model ${F}_{k}$ is divided into a personalized bottom model $\varphi_k$ and a global top model $H$. Specifically, each client $k$ maintains a composite model defined as $F_k(w_k) = \varphi_k(v_k) \circ H(\theta_k)$, where $v_k$ and $\theta_k$ represent the parameters of the bottom and top models, respectively. 

\begin{figure*}
    \centering
    \includegraphics[width=1\linewidth]{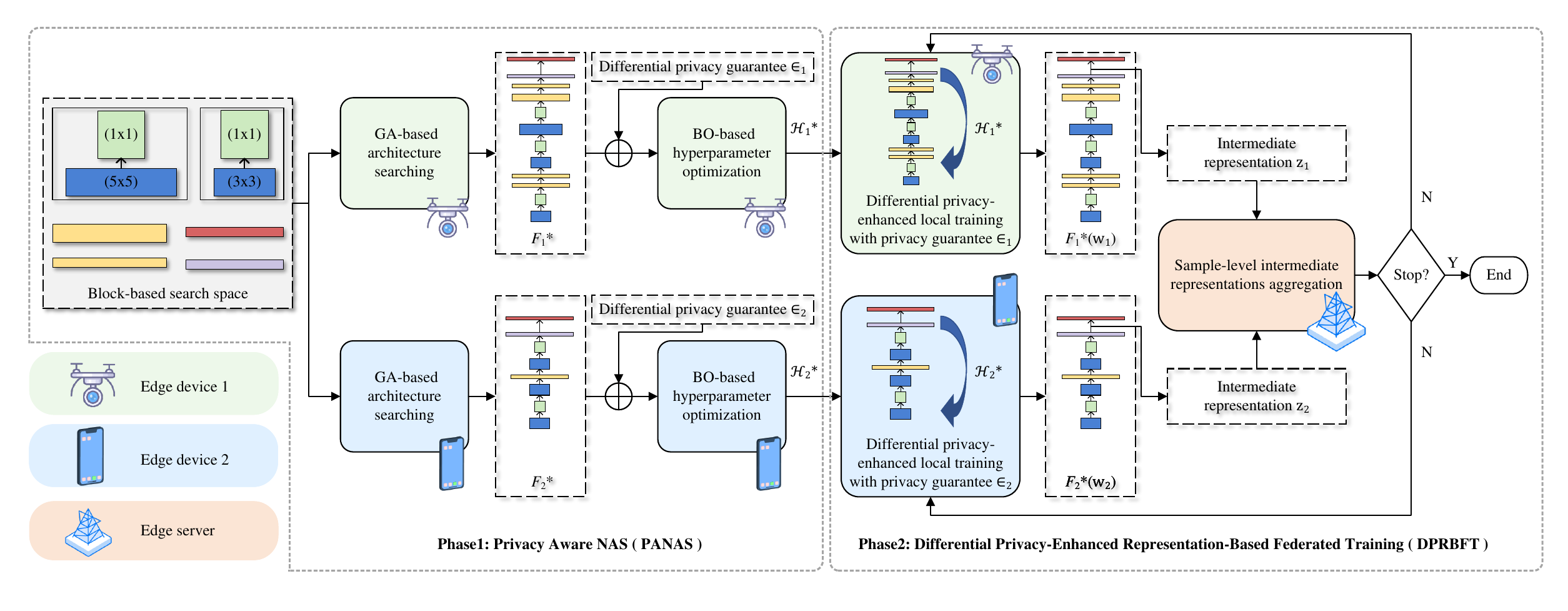}
    \caption{Framework of DPFNAS}
    \label{fig:DPFNASworkflow}
\end{figure*}

The Framework of the proposed Differential Privacy-Enhanced FNAS (DPFNAS) is illustrated in Fig.\ref{fig:DPFNASworkflow}, which mainly includes the following two phases. Firstly, we propose a Privacy Aware NAS (PANAS) algorithm to obtain an optimal model architecture $F_k^*=\varphi_k^* \circ  H $ and optimal hyperparameter configuration $\mathcal{H}_k^*$ tailored to each device's local data distribution under differential privacy guarantee $\epsilon_k$. Subsequently, we propose a Differential Privacy-Enhanced Representation-Based Federated Training (DPRBFT) strategy to mitigate the privacy risk of training data reconstruction during collaborative training.

\subsection{PANAS Algorithm}
\label{PANAS Algorithm}
By the proposed PANAS, N architectures are firstly explored within a novel lightweight block-based search space, forming a population where each architecture is considered an individual. This population is then evolved using the genetic algorithm (GA) -based architecture search mechanism to select the optimal architecture. Subsequently, the Bayesian optimization (BO)-based hyperparameter optimization strategy is employed to automatically identify the optimal configuration for the selected architecture under a fixed privacy budget. By executing the optimization process sequentially, the proposed PANAS ultimately derives both the optimal architecture and its corresponding hyperparameter configuration. The pseudo code of PANAS is presented in Algorithm \ref{alg:panas}, which works in detail as follows.

\begin{algorithm}[t]
\caption{PANAS:  Privacy-Aware NAS}
\label{alg:panas}
\begin{algorithmic}[1]
\State \textbf{Input:} \textbf{Block-based search space $\mathcal{S}$}, population size $N$, number of generations $T$, privacy budget $\epsilon$
\State \textbf{Output:} Optimal architecture $F^*$ and hyperparameter configuration $\mathcal{H}^*$

\State \textbf{// Phase 1: GA-Based Architecture Search}
\State Initialize population $\mathcal{P} = \{F_1, F_2, \dots, F_N\}$ by sampling $N$ architectures from $\mathcal{S}$
\For{$t = 1$ to $T$}
    \State Evaluate validation accuracy for each $F_i \in \mathcal{P}$
    \State Select parents using binary tournament selection
    \State Generate offspring via block-level crossover and mutation
    \State Apply environmental selection using roulette + elitism
    \State Update population $\mathcal{P}$
\EndFor
\State $F^* \gets \arg\max_{F \in \mathcal{P}} \text{ValAcc}(F)$

\vspace{1mm}
\State \textbf{// Phase 2: BO-Based Hyperparameter Optimization}
\State Define hyperparameter space $\Theta := \{\eta, B, C, \sigma\}$

\State Initialize candidate set $\mathcal{H}_0$ with $k$ random configurations $\theta_i \sim \Theta$
\For{each $\mathcal{H}_i \in \mathcal{H}_0$}
    \State Train $F^*$ using DP-SGD with $\mathcal{H}_i$, record $\text{ValAcc}(\mathcal{H}_i)$
    \State Compute $\text{PrivacyCost}(\mathcal{H}_i)$, discard if exceeds $\epsilon$
\EndFor
\State Fit GP surrogate model $f: \Theta \rightarrow \text{ValAcc}$ using $(\theta_i, \text{ValAcc}_i)$
\For{BO iteration $j = 1$ to $M$}
    \State Use Expected Improvement to select next $\mathcal{H}_j$
    \State Train $F^*$ with $\mathcal{H}_j$, evaluate $\text{ValAcc}(\mathcal{H}_j)$
    \State If $\text{PrivacyCost}(\mathcal{H}_j) \leq \epsilon$: update surrogate model $f$
\EndFor
\State $\mathcal{H}^* \gets \arg\max_{\mathcal{H} \in \Theta, \text{PrivacyCost}(\theta) \leq \epsilon} f(\mathcal{H})$
\State \Return $F^*$ and $\mathcal{H}^*$
\end{algorithmic}
\end{algorithm}

\subsubsection{\textbf{Block-Based Search Space Design}}
\begin{figure}
    \centering
    \includegraphics[width=1\linewidth]{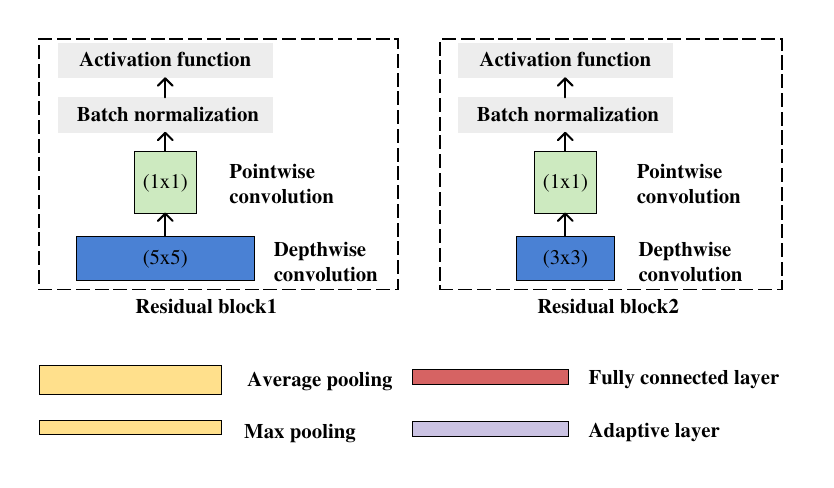}
    \caption{Design of the Block-Based Search Space}
    \label{fig:Block-baseddesign}
\end{figure}

The search space defines the set of all valid architectures, and NAS is essentially the process of searching for the optimal individual within this space \cite{liu2021survey}. Thus, the size of the search space can significantly impact the efficiency of the search process. The proposed block-based search space is illustrated in Fig.\ref{fig:Block-baseddesign}. To reduce the size of search space, we leverage the block-based search space design strategy \cite{lu2022evolving}, in which heterogeneous layers are grouped into reusable blocks that serve as the basic unit of architectural composition. Compared to the layer-based search space, where the basic units are primitive layers, the block-based search space makes it easier to discover promising architectures, and is therefore more suitable for resource-constrained edge devices in 6G networks.

The design of a block-based search space needs to consider both how blocks are composed with one another and how the layers within each block are structured. To enhance the flexibility of the search space, we adopt a variable-length block composition strategy. This means that the number of blocks in an architecture can be arbitrarily stacked during the search process, resulting in individuals with variable lengths. Compared to fixed-length strategies \cite{baldeon2020adaresu, chen2019auto}, the variable-length design has the advantage of not requiring additional expertise or prior experience to predefine an appropriate architecture length for each client. Instead, it allows the architecture to adaptively adjust its depth based on the data distribution during the search process. This flexible design is particularly well-suited for edge intelligence scenarios, where data distributions are highly diverse. It is worth noting that although our search space does not impose a strict constraint on architecture length, in practice, we usually define an upper and lower bound for the number of blocks. Such a constraint is essential for ensuring the efficiency of the NAS process. 

To ensure the lightweight nature of the search space, we adopt a compact residual block design. Specifically, each residual block consists of a Depthwise Separable Convolution\cite{chollet2017xception} (i.e. a Depthwise Convolution followed by a Pointwise Convolution), Batch Normalization \cite{santurkar2018does}, and a non-linear activation function (i.e. Relu) \cite{chen2020dynamic}, applied sequentially. Besides, a shortcut connection \cite{he2016deep} is added to facilitate stable gradient flow. Notably, the configuration of each layer within the block is also included in the search space. For instance, the kernel size of the Depthwise Convolution is searchable, with 3×3 and 5×5 as candidate options. The main advantage of Depthwise Separable Convolutions lies in their ability to dramatically reduce both the number of parameters and computational cost while maintaining strong representational power \cite{haase2020rethinking}. By combining Depthwise Separable Convolutions with residual connections, this block structure achieves significantly lower computational overhead and parameter count compared to standard convolutional blocks, while preserving effective feature transformation capabilities. 

Beyond convolutional blocks, we further enrich the search space by incorporating pooling operations, where both average pooling \cite{lecun2002gradient} and max pooling \cite{krizhevsky2012imagenet} are considered as searchable choices. Incorporating pooling layers enhances the model’s ability to perform spatial downsampling and control overfitting, while preserving essential features. By allowing the type of pooling operation to be searched, the architecture can better adapt to varying data characteristics and computational constraints \cite{bieder2021comparison}.

To ensure the searched architectures are deployable under the representation-based federated training paradigm, we introduce specific constraints into the search space design. To ensure global consistency, the "global head" of each architecture is fixed to a predefined shared structure. Additionally, to enable the aggregation of intermediate representations, an adaptation layer is appended as the final layer of each architecture’s feature extractor. This adaptation layer produces a fixed-dimensional output that matches the input dimension required by the global head.

\subsubsection{\textbf{GA-Based Architecture Search Mechanism}}
To efficiently search the optimal architecture within the proposed block-based search space, we design a GA-based architecture search mechanism. The mechanism is designed based on the general workflow of GA \cite{holland1992genetic}, which performs a series of evaluation, selection, crossover, and mutation operations to thoroughly explore diverse architectures in the search space. A key advantage of GA is its insensitivity to local minima, thus ensuring greater architectural variety \cite{alhijawi2024genetic}.

The overall workflow of the proposed GA-based architecture search mechanism is outlined in Algorithm \ref{alg:panas}. It begins by generating an initial population from the search space. Each individual in the population is then evaluated through a fitness evaluation process. Based on their fitness, the most promising individuals are selected via parent selection, and undergo crossover and mutation operations to produce a new set of offspring. Subsequently, an environmental selection process is applied to the combined pool of parents and offspring to select the best individuals for forming the new population. This new population then proceeds to the next iteration. The process repeats until a stopping criterion is met, at which point the optimal individual is output.

To adapt to the edge intelligence environment and the proposed block-based search space, we made the following design choices in the GA-based architecture search mechanism. Firstly, to minimize human intervention in the search process and better suit edge intelligence scenarios, we adopt a random space strategy \cite{sun2019evolving} during the population initialization step. And in the fitness evaluation step, validation accuracy is used as the evaluation metric. Considering the limited computational resources on edge devices, we incorporate an early stopping policy, which is widely used in NAS to reduce training cost. Specifically, we set a fixed and relatively small number of training epochs, a strategy that has been proven effective in saving computational resources \cite{liu2021survey}.

Although both parent selection and environmental selection are selection operations, their objectives differ, and thus we design distinct strategies for each. The goal of parent selection is to choose individuals as parents in order to generate offspring through crossover and mutation. To preferentially select individuals with higher accuracy, we adopt the binary tournament selection \cite{fang2010review} in the parent selection step. Specifically, two different individuals are randomly selected from the current population, and the one with the higher accuracy is chosen to participate in the variation process. This design is accuracy-driven but also introduces randomness to avoid premature convergence to local optima. In contrast, the goal of environmental selection is to form the new population for the next generation. To ensure broader exploration of the search space, it is important not only to retain high-accuracy individuals but also to preserve population diversity. Therefore, in the environmental selection step, we employ a combination of Roulette Wheel Selection (RWS) \cite{lipowski2012roulette} and an elitism preservation mechanism. The RWS assigns each individual a survival probability proportional to its accuracy, meaning that every individual, regardless of its fitness, has a chance to be eliminated. This helps maintain diversity in the population. Meanwhile, to ensure the efficiency of the search process, we incorporate elitism to guarantee that the best-performing individual is always retained in the next generation. Specifically, if the individual with the highest accuracy is not selected by the roulette mechanism, we replace the lowest-accuracy individual in the selected set with the best one.

Both the crossover and mutation steps generate new individuals by modifying the composition of existing ones, with the goal of exploiting high-performing individuals and exploring the broader search space. The key difference lies in that crossover operates between two individuals, whereas mutation is applied to a single individual. In our framework, both crossover and mutation are performed at the block level. Specifically, during crossover, two individuals are randomly selected from the parent population, and a random cut-point is selected for each. The structures of the two individuals are then swapped at the respective cut-points, resulting in two new individuals. For the resulting offspring, a mutation operation is applied based on a predefined probability. Specifically, we define four mutation strategies tailored to the block-based architecture representation. In the \textbf{Add} operation, a random block is inserted at a randomly selected position within the individual. The \textbf{Remove} operation deletes a randomly chosen block. The \textbf{Alter} operation selects a block at random and modifies the configuration of the layers within it. Finally, the \textbf{Exchange} operation randomly selects two blocks within an individual and swaps their positions. We introduce constraints in both crossover and mutation operations to ensure the validity of the generated individuals. In addition, we define the probability of performing crossover and mutation to control the balance between exploitation and exploration.

The termination criterion of the search is defined as a fixed number of generations, which is a widely adopted and effective stopping strategy in genetic algorithms.

\subsubsection{\textbf{BO-Based Hyperparameter Optimization Strategy}}

To ensure model utility under strict differential privacy constraints and mitigate the computational burden incurred by repeatedly training models during evaluation, we introduce the BO-based hyperparameter optimization strategy to address the privacy-constrained hyperparameter optimization problem.

To mitigate privacy leakage caused by intermediate representations, we adopt the DP-SGD \cite{abadi2016deep} during training. Models trained with DP-SGD exhibit heightened sensitivity to training hyperparameters compared to architectural choices~\cite{abadi2016deep}. We formalize the problem of ensuring model accuracy as a privacy-constrained hyperparameter optimization task:
\begin{multline}
    \mathcal{H}^* = \arg\max_{\mathcal{H}} \, \text{ValAcc}(\mathcal{H}) \\
    \text{s.t.} \quad \text{PrivacyCost}(\mathcal{H}) \leq \text{PrivacyBudget}
\end{multline}
Here, the hyperparameter configuration is defined as $\mathcal{H} := {\eta, B, C, \sigma}$, where $\eta$ is the learning rate, $B$ is the batch size, $C$ is the clipping threshold, and $\sigma$ is the noise scale. Among these, $C$ and $\sigma$ are privacy-related hyperparameters. DP-SGD enforces differential privacy by clipping the $\ell_2$-norm of each individual gradient to a predefined threshold $C$, and then adding Gaussian noise with standard deviation $\sigma$ to the clipped gradients.

Solving the DP-SGD hyperparameter optimization problem on resource-constrained edge devices is particularly challenging, as the process requires evaluating numerous hyperparameter configurations through training, and training with DP-SGD is known to be computationally expensive \cite{ponomareva2023dp}. The BO strategy \cite{vincent2023improved} constructs a surrogate model to approximate the relationship between hyperparameter configurations and model performance, thereby avoiding the need for full training during each evaluation. 

The workflow of our proposed BO-based hyperparameter optimization strategy is presented in Algorithm \ref{alg:panas}. Firstly, a small subset of candidate hyperparameter configurations are randomly sampled from the overall search space. Each configuration in this subset is then used to train a model and obtain the corresponding validation accuracy, while ensuring that the privacy cost remains within the predefined privacy budget. Based on these observed evaluations, a surrogate model is constructed to approximate the performance landscape. Given that our search space involves only four hyperparameters, we adopt a Gaussian Process (GP) \cite{ko2009gp} as the surrogate model, as it is well-suited for low-dimensional, constraint-aware optimization scenarios \cite{liu2020gaussian}. Despite being built on a limited number of observed evaluations, the GP can provide a probabilistic estimation of validation accuracy across the entire hyperparameter space. Next, an acquisition function (AF) is employed to select the next hyperparameter configuration to evaluate. The selected configuration is expected to either improve upon the current best validation accuracy or reduce uncertainty in poorly explored regions. To effectively balance exploration and exploitation, we adopt the Expected Improvement (EI) \cite{zhan2020expected} acquisition function, which favors configurations that are likely to yield performance gains while also accounting for model uncertainty. This iterative process continues until a predefined number of rounds is reached. The configuration with the highest surrogate-predicted validation accuracy under the differential privacy constraint is then selected as the optimal hyperparameter configuration.

\subsection{DPRBFT strategy}

\begin{figure}
    \centering
    \includegraphics[width=1\linewidth]{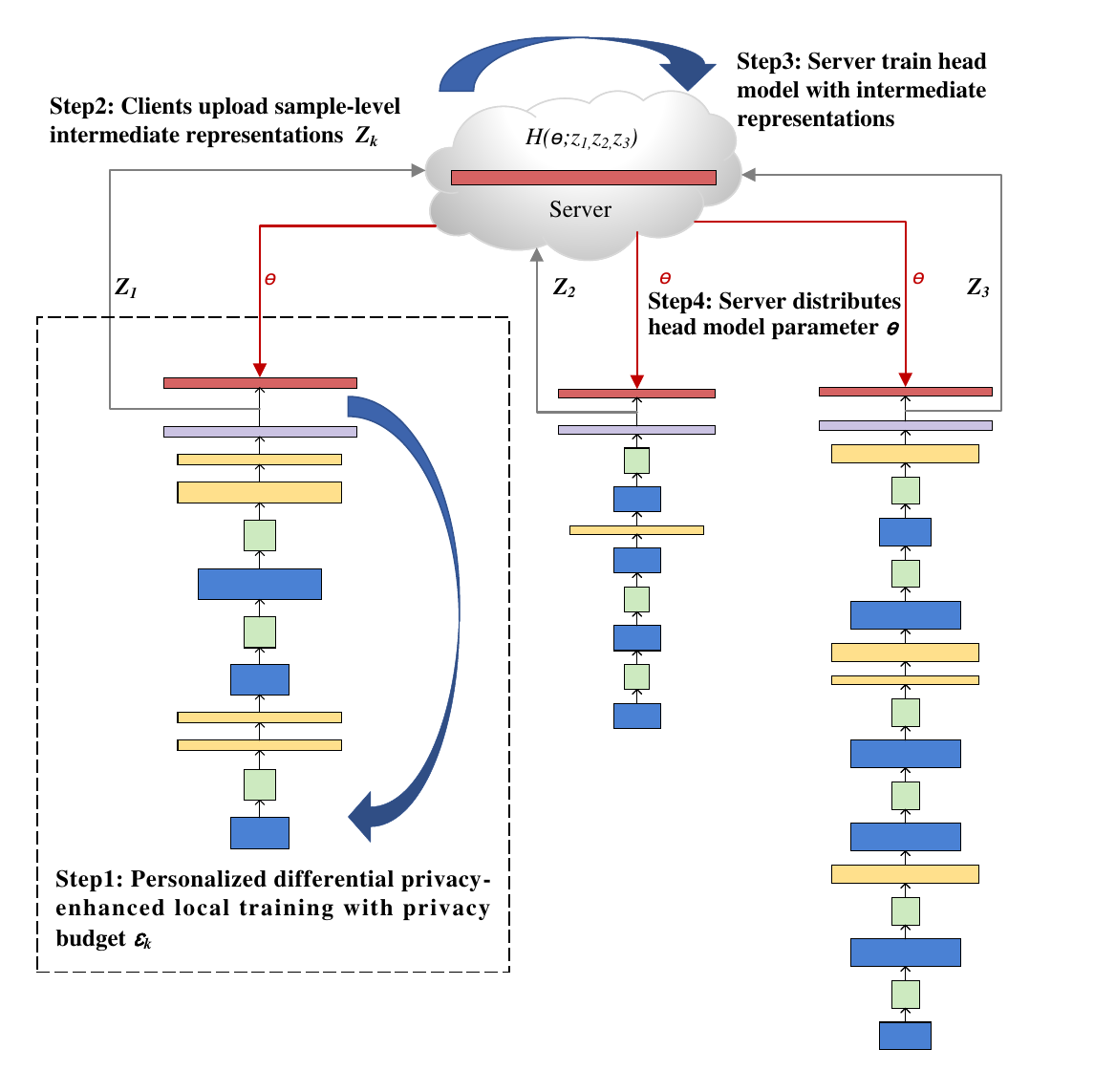}
    \caption{Workflow of the DPRBFT strategy}
    \label{fig:DPRBFTframework}
\end{figure}

To support privacy-preserving collaborative training for heterogeneous optimal models, we propose the DPRBFT strategy. By the proposed DPRBFT, each client first performs local training in parallel using the architecture and hyperparameters derived from the proposed PANAS. During this phase, a personalized differential privacy-enhanced strategy is applied to protect against potential reconstruction attacks from intermediate representations. After local training, the clients perform knowledge transfer via a sample-level intermediate representation aggregation strategy, which enables robust collaboration among heterogeneous local models. This local training and global aggregation process is repeated iteratively until a predefined stopping criterion is met. Ultimately, each client obtains a fully trained model. The pseudo code of DPRBFT is presented in Algorithm \ref{alg:dprbft} and the workflow is illustrated in Fig.\ref{fig:DPRBFTframework}. 

\begin{algorithm}[t]
\caption{DPRBFT: DP-Enhanced Representation-Based Federated Training}
\label{alg:dprbft}
\begin{algorithmic}[1]
\State \textbf{Input:} Optimal architecture $F_k^*$, optimal hyperparameters $\mathcal{H}_k^*$ for each client $k$ from PANAS; number of communication rounds $T$
\State \textbf{Output:} Trained personalized model for each client $F_k^* = \varphi_k^*(v_k^*) \circ H(\theta^*)$

\For{each client $k \in \{1, \dots, N\}$}
    \State Initialize local model: $F_k^*(w_k) = \varphi_k^*(v_k) \circ H(\theta_k)$
\EndFor

\For{round $t = 1$ to $T$}
    \For{each client $k$ in parallel}
        \State Train local model $F_k^*(w_k)$ using DP-SGD with $\mathcal{H}_k^*$
        \State Compute intermediate representations: $z_k = \varphi_k^*(v_k; x_k)$
        \State Send $z_k$ and corresponding labels $y_k$ to server
    \EndFor
    \State Server aggregates $\{(z_k, y_k)\}$ and updates global top model:
    \[
        \theta \leftarrow \arg\min_{\theta} \sum_{k=1}^{N} \frac{m_k}{m} \mathcal{L}(H(\theta; z_k), y_k)
    \]
    \State Server broadcasts updated $\theta$ to all clients
    \For{each client $k$}
        \State Update local top model: $\theta_k \leftarrow \theta$
    \EndFor
\EndFor

\State \Return $F_k^*(w_k) = \varphi_k^*(v_k^*) \circ H(\theta^*)$ for all $k$
\end{algorithmic}
\end{algorithm}

\subsubsection{\textbf{Personalized Differential privacy-enhanced local training}}
To mitigate potential privacy leakage caused by intermediate representations in representation-based collaborative training, we adopt a personalized, differential privacy-enhanced local training strategy. Based on the optimal model architecture $F_k^*$ and optimal hyperparameter configuration $\mathcal{H}_k^*$ obtain from the PANAS, each client $k$ initializes and maintains a local model defined as $F_k^*(w_k) = \varphi_k^*(v_k) \circ H(\theta_k)$. During each round, clients first perform local training on their private data using DP-SGD \cite{abadi2016deep} with their personalized hyperparameter configurations $\mathcal{H}_k^*$. 

To accommodate the personalized privacy requirements of different clients, our method allows each client to specify a distinct privacy budget. This personalized setting still adheres to rigorous differential privacy guarantees. In our design, differential privacy is enforced only during the local training phase on each client using DP-SGD. Each client independently consumes its own privacy budget according to its local hypermeter configuration. After local training, only intermediate representations are uploaded to the server. Based on the post-processing immunity property \cite{dwork2006differential} of differential privacy, these features are generated by models already trained with differential privacy guarantees, and therefore do not incur any additional privacy cost. That is, any function applied to the output of a DP mechanism remains differentially private, with the same privacy parameters, as long as no further access to the original private data is involved. The server-side training of the shared classifier head is performed without access to raw data or gradients and thus does not contribute to any client’s privacy budget consumption. Hence, the total privacy cost for each client is solely determined by its local DP-SGD training process. 

It is also worth noting that, to the best of our knowledge, our work is the first one to apply the representation-based FL paradigm with personalized DP-SGD. Given that attackers can easily reconstruct training data from intermediate representations~\cite{pasquini2021unleashing}, this design choice is both necessary and well-justified.

\subsubsection{\textbf{Sample-level intermediate representations aggregation}}
Upon completing local training, clients upload intermediate representations $z_k = \varphi_k^*(v_k; x)$ to the server, which uses them as inputs for optimizing the global top model $H$. The global optimization objective is formulated as:

\begin{equation}
    \arg\min_{\theta} \sum_{k=1}^{N} \frac{m_k}{m} \mathcal{L}(H(\theta; z_k), y),
\end{equation}

where $m_k$ is the size of client $k$’s dataset and $m = \sum_k m_k$. 

Unlike existing representation-based FL approaches such as FedProto~\cite{tan2022fedproto}, FedClassAvg~\cite{jang2023fedclassavg}, and FedGH~\cite{yi2023fedgh}, which construct class prototypes by averaging feature representations within each class—an approach that assumes balanced class distributions—our method uploads per-sample representations without class-wise aggregation.
We intentionally choose to upload individual sample representations without averaging, which mitigates the adverse effects of label imbalance and better captures the non-IID nature of data on edge devices.
The results in Experiment~\ref{Robustness to System Heterogeneity} demonstrate the robustness of this design.

After optimizing the global model $H$, the server broadcasts the updated parameters $\theta$ to all clients, who then update their local top models to match the global version, i.e., $\theta_k \leftarrow \theta$. This completes one round of federated training.

\subsection{Convergence Analysis}

In this subsection, we conduct a convergence analysis for DPFNAS. We make the following assumptions similar to existing general representation-based federated training approaches \cite{yi2023fedgh, tan2022fedproto}.

\begin{assumption}
    For any client \( k \) and \( w \in \mathbb{R}^d \), the gradient of the local loss function is bounded:
    \[
    \|\nabla \mathcal{L}_k(w)\| \leq B
    \]
\end{assumption}

\begin{assumption}
    The local loss function is $L$-smooth:
    \[
    \|\nabla \mathcal{L}_k(w) - \nabla \mathcal{L}_k(w')\| \leq L \|w - w'\|
    \]
\end{assumption}

\begin{assumption}
    The deviation between the local prediction head \( H_k(\theta_k) \) and the global prediction head \( H(\theta) \), due to partial training and personalization, is bounded by:
    \[
    \mathbb{E}[\|\theta_k - \theta\|^2] \leq \alpha^2
    \]
\end{assumption}

\begin{assumption}
    The stochastic gradient \( g_{k,t} \) is an unbiased estimator of the full local gradient:
    \[
    \mathbb{E}_{\xi_k \sim \mathcal{D}_k}[g_{k,t}] = \nabla \mathcal{L}_k(w_{k,t}), \quad \forall k
    \]
    with bounded variance:
    \[
    \mathbb{E}[\|g_{k,t} - \nabla \mathcal{L}_k(w_{k,t})\|_2^2] \leq \sigma^2
    \]
\end{assumption}

\begin{assumption}
    The gradient of the shared top model satisfies:
    \[
    \mathbb{E}[g_{\theta,t}] = \nabla \mathcal{L}_\theta(\theta_t)
    \]
\end{assumption}

\begin{theorem}[Local Model Loss Bound within One Communication Round]
Under the above assumptions, the expected loss of the local model after one communication round is bounded by:
\begin{align}
\mathbb{E}[\mathcal{L}_{k}(w_{tE+E})]\leq & \, \mathcal{L}_{k}(w_{tE+0}) \nonumber \\
& -(\eta p-\frac{L\eta^{2}}{2})\sum_{e=0}^{E-1}\|\nabla\mathcal{L}_{k}(w_{tE+e})\|^{2} \nonumber \\
& +\frac{EL\eta^{2}}{2}(\sigma^{2}+d \cdot \delta^{2} C^{2})
\end{align}
where $tE$ denotes the local model before fine-tuning, $E$ is the number of local epochs, and $p$ is a constant dependent on clipping norm $C$ and gradient bound $B$.
\end{theorem}

\begin{corollary}[Local Model Loss Bound After Aggregation]
\begin{align}
\mathbb{E}[\mathcal{L}_k (w_{tE+E+0})] \leq & \, \mathcal{L}_k (w_{tE+0})\nonumber \\
& - \left( \eta_w p - \frac{L\eta_w^2}{2} \right) \sum_{e=0}^{E-1} \|\nabla \mathcal{L}_k (w_{tE+e})\|^2 \nonumber \\
& + \frac{E L \eta_w^2}{2} (\sigma^2 + d \cdot \delta^2 C^2) + \eta_w E B^2\nonumber \\
& + \eta_\theta B + \frac{L}{2} \alpha^2
\end{align}
\end{corollary}

\begin{corollary}[Convergence Bound After \( T \) Training Rounds]
\begin{align}
\frac{1}{T} \sum_{t=0}^{T-1} \sum_{e=0}^{E-1} \|\nabla \mathcal{L}_k (w_{tE+e})\|^2 \leq & \, \frac{1}{\eta_w p - \frac{L\eta_w^2}{2}} \nonumber \\
& \bigg( \frac{\Delta}{T} + \frac{E L \eta_w^2}{2} (\sigma^2 + d \cdot \delta^2 C^2) \nonumber \\
& + \eta_w E B^2 + \eta_\theta B + \frac{L}{2} \alpha^2 \bigg)
\end{align}
\end{corollary}


To ensure convergence, the local learning rate must satisfy:
\begin{align}
\eta_w \leq & \, 
\frac{ pG - EB^{2}}{L \big( G + E (\sigma^{2} + d \cdot \delta^{2} C^{2}) \big) B} \nonumber \\
& + \frac{ \sqrt{(pG - EB^{2})^{2} 
+ 2L \big( G + E (\sigma^{2} + d \cdot \delta^{2} C^{2}) \big) }  
}{L \big( G + E (\sigma^{2} + d \cdot \delta^{2} C^{2}) \big) B} \nonumber \\
& \quad 
+ \frac{ 
\left(\eta_{\theta} B + \frac{L}{2} \alpha^{2} \right) 
}{L \big( G + E (\sigma^{2} + d \cdot \delta^{2} C^{2}) \big) B}
\end{align}

Global learning rate:
\[
0 \leq \eta_\theta \leq \frac{\alpha - \eta_w E B}{B}
\]

The above theoretical results establish that under standard assumptions commonly adopted in federated optimization, the proposed DPFNAS framework achieves convergence in expectation. Specifically, \textbf{Theorem 1} provides a per-round local loss bound that captures the impact of stochastic gradients, privacy-induced noise, and local updates. \textbf{Corollary 1} extends this result to include the effects of global aggregation and local prediction head personalization, where the deviation between local and global heads is upper bounded by a constant $\alpha$, i.e., $\mathbb{E}[\|\theta_k - \theta\|^2] \leq \alpha^2$. \textbf{Corollary 2} further demonstrates that the average squared gradient norm over $T$ rounds diminishes as $T$ increases, indicating convergence to a stationary point. A detailed proof and analysis are given in Appendix.

These results confirm that the proposed DPFNAS maintains convergence even under differential privacy constraints, owing to the proper calibration of local and global learning rates $\eta_w$ and $\eta_\theta$ as bounded in the final inequality. The convergence bound also reveals the trade-off between privacy noise and convergence rate, which can be managed by tuning the learning rate, batch size, clipping tehreshold and noise scale. This finding further offers theoretical support for the design of our BO-based hyperparameter optimization strategy in \ref{PANAS Algorithm}.

In summary, the proposed DPFNAS provides a provably convergent training framework that accommodates personalization, privacy noise, and statistical heterogeneity—making it theoretically sound and practically robust for federated learning in privacy-critical scenarios.

\section{Experiments}

In this section, we conduct a comprehensive experimental evaluation of the proposed DPFNAS framework. The experiments are designed to validate its performance from multiple perspectives. Specifically, we first compare the proposed DPFNAS with state-of-the-art structure-agnostic \cite{yi2023fedgh,wang2023flexifed} and NAS-based federated learning methods \cite{yao2024perfedrlnas} to assess its overall effectiveness in terms of accuracy, convergence speed, model compactness, and communication efficiency. We then evaluate the privacy-preserving capability of the proposed DPFNAS through feature inversion attacks, examining reconstruction quality under varying privacy guarantees. Furthermore, we perform ablation studies to analyze the robustness of the proposed sample-level intermediate representation aggregation strategy, the flexibility of the block-based search space and GA-based architecture search mechanism, and the effectiveness of the BO-based hyperparameter optimization strategy.

\subsection{Experimental Setup}

\subsubsection{\textbf{Datasets and Heterogeneity Simulation}}

All experiments are performed on the CIFAR-10 and CIFAR-100 datasets. To simulate realistic statistical heterogeneity commonly observed in federated learning scenarios, we adopt the non-IID partitioning strategy proposed in FedGH~\cite{yi2023fedgh}. Specifically, each client is assigned data from a fixed subset of classes with non-uniform sampling probabilities, effectively emulating the non-identical data distributions found in practical deployments.

For baseline methods that do not incorporate NAS, heterogeneous Convolutional Neural Network (CNN) \cite{lecun2002gradient} architectures are randomly assigned to clients according to the configuration scheme used in FedGH\cite{yi2023fedgh}. These five CNN variants differ in the width of convolutional filters and the size of fully connected layers, thereby introducing structural heterogeneity across clients.

\subsubsection{\textbf{Baselines for Comparison}}  
We compare the proposed DPFNAS with several state-of-the-art personalized and heterogeneous federated learning baselines:

\begin{itemize}
    \item \textbf{FedGH}~\cite{yi2023fedgh}: Enables the aggregation of fully heterogeneous models through local alignment regularization, improving consistency across clients without requiring shared architectures.
    
    \item \textbf{PerFedRLNAS}~\cite{yao2024perfedrlnas}: A reinforcement learning-based NAS framework designed for federated settings that adapts model architectures on the client side based on resource constraints and data characteristics.
    
    \item \textbf{FlexiFed}~\cite{wang2023flexifed}: Supports model heterogeneity within a predefined architecture family, allowing flexible client updates while preserving aggregation compatibility.
\end{itemize}

In addition, we include \textit{local training} (i.e., standalone learning without collaboration) as a baseline to highlight the benefits of federated collaboration under non-IID data conditions.

\subsubsection{\textbf{Training Configuration}}  
 
All methods evaluated in our experiments—including our proposed DPFNAS and all baselines—adopt stochastic gradient descent (SGD) \cite{amari1993backpropagation} as the local optimizer. The learning rate is fixed at 0.01, a commonly adopted setting in federated optimization and validated through our preliminary experiments. To ensure fairness, the number of local training epochs is set to 30 and kept consistent across all methods. The batch size is selected from {64, 128, 256, 512} via validation tuning and then fixed for each method. To ensure that all methods are trained for sufficient rounds to reach convergence, the total number of communication rounds is set to 300. It is important to note that, to ensure a fair comparison of model performance, our proposed method does not apply DP-SGD during training and does not use the BO-based hyperparameter optimization strategy in any of the experiments presented in Section~\ref{sec:Comparison with Structure-Agnostic FL Methods} and \ref{sec:Comparison with NAS-based FL Methods}.

In the experiments presented in Section~\ref{sec:Comparison with Structure-Agnostic FL Methods} and Section~\ref{sec:Ablation Study}, we set the number of participating clients to 10, with all clients involved in every communication round. This setting is chosen to ensure a fair comparison, following the same experimental configuration as FedGH~\cite{yi2023fedgh}, which also evaluates performance under heterogeneous model settings with 10 clients. In the experiments in Section~\ref{sec:Comparison with NAS-based FL Methods}, we set the number of participating clients to 100, with a participation rate of 10\% per communication round. This configuration aligns with practical requirements outlined in the 6G Use Case Research report~\cite{IMT2030_6GUseCase_2024}, which indicates that collaborative training scenarios in edge intelligence typically involve hundreds of devices.

\subsubsection{\textbf{NAS Configuration}}  

Following the experimental configurations used in prior works\cite{lu2022evolving} on evolutionary neural architecture search, we adopt a lightweight NAS configuration with a small population size of 10 and a maximum of 20 generations. This setting is designed to more realistically simulate the limited storage and computational resources typical of edge devices. To further align with realistic scenarios, the local dataset on each client is explicitly partitioned into separate subsets for NAS and for subsequent model training. Each client performs NAS independently using a locally held data subset, which includes 500 training samples, 100 validation samples, and 100 test samples.

\subsubsection{\textbf{Evaluation Metrics}}  
We evaluate each method using the average test accuracy across all clients and the corresponding standard deviation, which reflects the level of personalization and inter-client consistency. The communication cost is derived in megabytes (MB) per client, representing the volume of exchanges of model parameters during training. Model size is calculated based on the total memory footprint of all trainable parameters.

\subsection{Overall effectivness of DPFNAS}
\label{sec:DPFNAS}
\subsubsection{\textbf{Comparison with Structure-Agnostic FL Methods}}  
\label{sec:Comparison with Structure-Agnostic FL Methods}

We first compare the proposed DPFNAS with two structure-agnostic baselines: FedGH \cite{yi2023fedgh}, a fully heterogeneous model aggregation framework, and Local Training, where each client trains its model independently without collaboration. In addition, we include DPFNAS (no NAS) as an ablation variant of our method, where the final training phase is retained but the architecture search stage is removed.
Experiments are conducted on CIFAR-10 and CIFAR-100 \cite{krizhevsky2009learning}, with target accuracy thresholds set to 93\% and 75\%, respectively. The results are summarized in Table~\ref{tab:Performance_comparison}.

\begin{table}[t!]
\caption{Comparison with structure-agnostic methods.}
\centering
\resizebox{\columnwidth}{!}{
\begin{tabular}{lrrrr}
\toprule
\multirow{2}{*}{\textbf{Method}} & \multicolumn{2}{c}{CIFAR-10} & \multicolumn{2}{c}{CIFAR-100} \\ 
\cmidrule(lr){2-3} \cmidrule(lr){4-5} 
 & \multicolumn{1}{r}{Acc (\%)} & \multicolumn{1}{r}{Rounds} & \multicolumn{1}{r}{Acc (\%)} & \multicolumn{1}{r}{Rounds} \\ 
\midrule
Local Training & 94.28 $\pm$ 2.81 & 5 & 71.94 $\pm$ 6.06 & - \\ 
FedGH \cite{yi2023fedgh} & 93.79 $\pm$ 3.07 & 13 & 73.29 $\pm$ 4.81 & - \\ 
DPFNAS (no NAS) & 93.82 $\pm$ 3.01 & 13 & 75.46 $\pm$ 4.75 & 88 \\ 
DPFNAS & \textbf{94.46 $\pm$ 2.30} & \textbf{4} & \textbf{76.82 $\pm$ 3.91} & \textbf{15} \\ 
\bottomrule
\end{tabular}
}
\label{tab:Performance_comparison}
\end{table}

\textbf{Accuracy:} As shown in Table~\ref{tab:Performance_comparison}, the full version of our method, DPFNAS, consistently achieves the highest test accuracy across both datasets. On CIFAR-10, the proposed DPFNAS reaches 94.46\% accuracy, outperforming FedGH \cite{yi2023fedgh} by +0.67\% and its own ablated variant DPFNAS (no NAS) by +0.64\%. This performance gain highlights the effectiveness of the NAS-based personalization module in improving model generalization.
On CIFAR-100, the accuracy advantage is even more substantial: the proposed DPFNAS achieves 76.82\%, exceeding FedGH \cite{yi2023fedgh} by +3.53\%. This suggests that the proposed DPFNAS is particularly robust in more challenging and fine-grained classification tasks. Furthermore, the proposed DPFNAS also exhibits a smaller standard deviation across clients, indicating better personalization and more stable performance under heterogeneous conditions.

\textbf{Convergence Speed:} In terms of communication efficiency, our proposed DPFNAS achieves the target accuracy significantly faster than its counterparts. On CIFAR-10, our proposed DPFNAS achieves the target 93\% accuracy in only 4 communication rounds, compared to 13 rounds required by both FedGH \cite{yi2023fedgh} and our proposed DPFNAS without NAS. On CIFAR-100, our proposed DPFNAS achieves the target 75\% accuracy in just 15 rounds, while the no-NAS variant requires 88 rounds, resulting in nearly a 6× speedup. These results highlight the dual advantage of the proposed framework in terms of both accuracy and training efficiency.

\textbf{Insights:} The comparison with local training provides additional insight into the benefit of collaborative learning. Although local training achieves high accuracy on CIFAR-10, it fails to generalize well on the more complex CIFAR-100 dataset. In contrast, our proposed DPFNAS not only maintains high accuracy, but also demonstrates the added value of collaboration, especially under statistically heterogeneous data and low-resource conditions.

\subsubsection{\textbf{Comparison with NAS-based FL Methods}} 
\label{sec:Comparison with NAS-based FL Methods}
To further evaluate the effectiveness of our proposed DPFNAS in architecture optimization under heterogeneous federated settings, we compare it with PerFedRLNAS\cite{yao2024perfedrlnas}, a state-of-the-art NAS-based personalized FL framework. The experiments are carried out on the CIFAR-100 dataset using a Dirichlet distribution with $\alpha = 0.1$ to simulate severe statistical heterogeneity. 
Two search spaces from PerFedRLNAS \cite{yao2024perfedrlnas} are evaluated: MobileNetV3 \cite{howard2019searching} and ViT \cite{dosovitskiy2020image}.

\textbf{NAS Efficiency:}
Table~\ref{tab:compare with pernas} presents the test accuracy and the resulting model size for each method. Our proposed DPFNAS achieves the highest accuracy of 47.28\%, outperforming both PerFedRLNAS-MobileNetV3 (+6.82\%) and PerFedRLNAS-ViT (+1.16\%). In particular, our proposed DPFNAS produces these results with a model size of only 1.17 MB—approximately 9.7× smaller than MobileNetV3 and 46.2× smaller than ViT. This demonstrates that our proposed lightweight block-based search space not only improves accuracy but also yields highly compact and deployment-friendly models.

\begin{table}[t!]
\caption{Comparison with federated NAS methods on CIFAR-100.}
\resizebox{\linewidth}{!}{
\begin{tabular}{@{}lrr@{}}
\toprule
\textbf{Method} & Accuracy (\%) & Model Size (MB) \\ 
\midrule
PerFedRLNAS-MobileNetV3 \cite{yao2024perfedrlnas} & 40.46 $\pm$ 6.44 & 10.87 $\pm$ 1.45 \\
PerFedRLNAS-ViT \cite{yao2024perfedrlnas} & 46.12 $\pm$ 5.39 & 54.04 $\pm$ 4.83 \\
DPFNAS & \textbf{47.28 $\pm$ 6.45} & \textbf{1.17 $\pm$ 0.47} \\
\bottomrule
\end{tabular}
}
\label{tab:compare with pernas}
\end{table}

\textbf{Communication Efficiency:}  
Beyond accuracy and model compactness, our proposed DPFNAS exhibits significant advantages in communication efficiency. As shown in Table~\ref{tab:trafficload_comparison}, we measure the communication overhead required to reach three accuracy thresholds (30\%, 35\%, and 40\%). 

Our proposed DPFNAS dramatically reduces the communication overhead across all target accuracy levels. For example, to reach 30\% accuracy, PerFedRLNAS-MobileNetV3 incurs a communication cost of 21.08 MB per client, whereas the proposed DPFNAS requires only 1.14 MB—a reduction by a factor of over 18.
Compared with PerFedRLNAS-ViT, the improvement is even more significant: the proposed DPFNAS lowers the communication overhead by approximately 94.8× at the 30\% target (108.08 MB vs. 1.14 MB), and achieves a savings of more than 800 MB per client at the 40\% target. 

\begin{table}[t!]
\caption{Communication overhead (MB per client) to reach target accuracy.}
\centering
\resizebox{\columnwidth}{!}{
\begin{tabular}{lrrr}
\toprule
\textbf{Method} & \textbf{30\%} & \textbf{35\%} & \textbf{40\%} \\ 
\midrule
PerFedRLNAS-MobileNetV3 \cite{yao2024perfedrlnas} & 21.08 & - & - \\ 
PerFedRLNAS-ViT \cite{yao2024perfedrlnas} & 108.08 & 540.40 & 972.72 \\ 
DPFNAS & \textbf{1.14} & \textbf{5.70} & \textbf{152.76} \\ 
\bottomrule
\end{tabular}
}
\label{tab:trafficload_comparison}
\end{table}

\textbf{Insights:}  
The superior performance of our proposed DPFNAS stems from the following three design advantages. Firstly, the use of lightweight building blocks ensures compact models that are better suited for low-resource clients. Secondly, its block-based modular search space simplifies architecture exploration, enabling faster convergence and more stable training. Thirdly, the decoupled representation-based design minimizes the need for synchronization, making our proposed DPFNAS more communication-efficient than centralized supernet-based NAS approaches like PerFedRLNAS. Overall, these results confirm that our proposed DPFNAS achieves a better trade-off among accuracy, model size, and communication overhead in federated NAS settings.

\subsubsection{\textbf{Privacy Protection Effectiveness}}

In this section, we evaluate the defense capability of the the proposed DPRBFT against feature inversion attacks under varying privacy budgets. 
We simulate a Feature Space Hijacking Attack (FSHA) \cite{pasquini2021unleashing}, where an adversarial server attempts to reconstruct input samples from the intermediate feature representations shared during training. The success of the attack is quantified using Mean Squared Error (MSE), where a lower value indicates more accurate reconstructions and thus greater privacy leakage.

Table~\ref{tab:privacy_protection} presents qualitative and quantitative results in varying privacy budgets $\epsilon$. Lower values of $\epsilon$ correspond to stronger privacy guarantees, as indicated by higher reconstruction error and reduced image clarity. Without any DP protection (i.e., $\epsilon = \infty$), the reconstructed images exhibit almost perfect fidelity to the original inputs (MSE = 0.02), clearly revealing sensitive client data. When a moderate privacy budget of $\epsilon = 50$ is applied, the quality of the reconstruction deteriorates visibly (MSE = 0.38) and critical visual details begin to vanish. At a strict privacy budget of $\epsilon = 5.0$ or below, the reconstructions become highly distorted (MSE $\geq$ 0.67), rendering them indistinguishable and effectively nullifying the effort of the attacker.

These findings confirm that the proposed DPRBFT provides a strong defense against feature-space inversion, especially in representation-based FL settings where intermediate representations are exchanged. The progressive degradation of reconstruction quality with decreasing $\epsilon$ further validates the controllability and effectiveness of our privacy-aware optimization.

\begin{table}[t!]
\caption{Reconstruction results under FSHA\cite{pasquini2021unleashing} at different privacy budgets ($\epsilon$).}
\centering
\resizebox{\columnwidth}{!}{
\begin{tabular}{>{\raggedleft\arraybackslash}p{1.2cm} >{\raggedleft\arraybackslash}p{1.5cm} >{\centering\arraybackslash}m{0.5\linewidth}}
\toprule
\textbf{$\epsilon$} & \textbf{MSE} & \textbf{Training Samples} \\ 
\midrule
\multicolumn{2}{c}{\textbf{Original Pictures}} & 
\includegraphics[width=\linewidth]{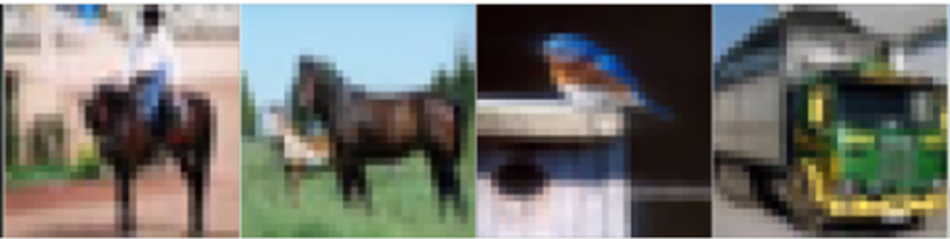} \\ 
\midrule
\textbf{Infinite} & 0.02 & 
\includegraphics[width=\linewidth]{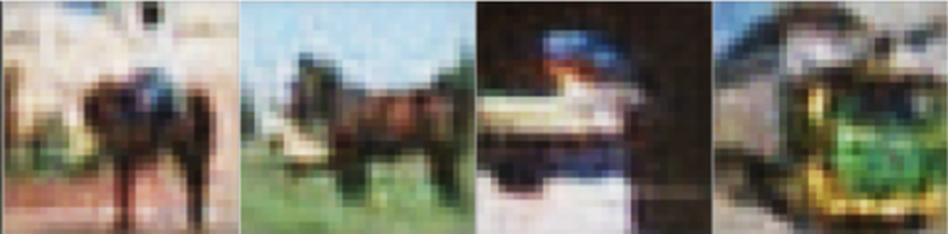} \\ 
\midrule
\textbf{50.00}    & 0.38 & 
\includegraphics[width=\linewidth]{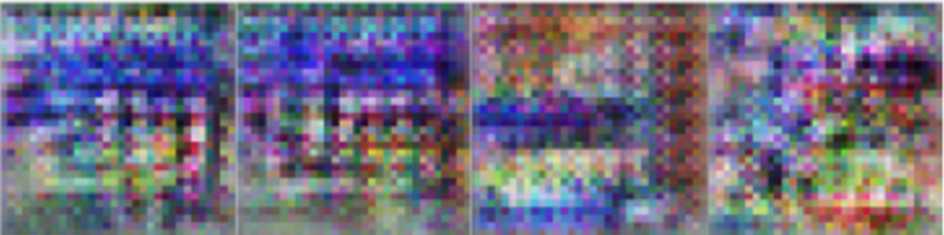} \\ 
\midrule
\textbf{5.00}     & 0.67 & 
\includegraphics[width=\linewidth]{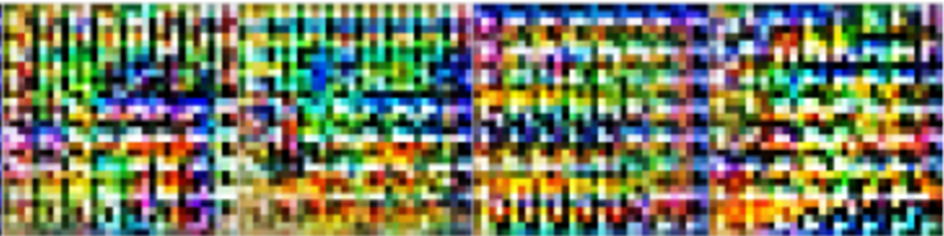} \\ 
\midrule
\textbf{0.50}     & 0.69 & 
\includegraphics[width=\linewidth]{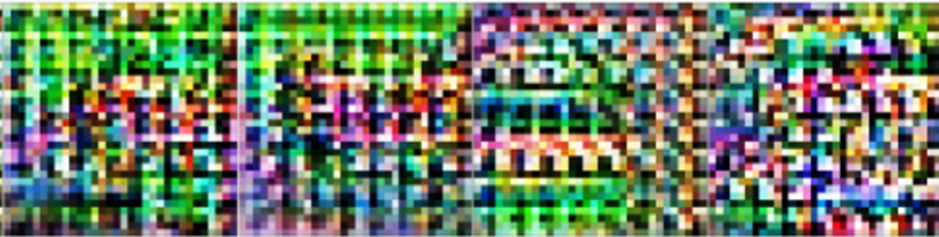} \\ 
\bottomrule
\end{tabular}
}

\label{tab:privacy_protection}
\end{table}
\subsection{Ablation Study}
\label{sec:Ablation Study}

\subsubsection{\textbf{Robustness of sample-level intermediate representations aggregation design}}  
\label{Robustness to System Heterogeneity}
To evaluate the robustness of our proposed sample-level intermediate representations aggregation method under system heterogeneity, we assign different model families to clients, including CNNs \cite{lecun2002gradient}, MobileNets \cite{howard2019searching} , and ResNets \cite{he2016deep}. This simulates a practical federated environment where devices have diverse computational capabilities and architectural preferences.

Table~\ref{tab:accuracy_comparison} reports the average precision between clients using each model family. Our proposed DPFNAS consistently achieves the highest precision in all groups - 79. 29\% for CNNs, 73. 68\% for MobileNets and 77. 84\% for ResNets - outperforming FlexiFed \cite{wang2023flexifed} and FedGH \cite{yi2023fedgh} by margins of up to 3.36\%. This indicates that the architecture search strategy for our proposed DPFNAS can effectively tailor models to diverse structural constraints, leading to robust and generalizable performance across heterogeneous deployments.

\begin{table}[t!]
\caption{Accuracy comparison across different model families.}
\centering
\resizebox{\columnwidth}{!}{
\begin{tabular}{lrrr}
\toprule
\textbf{Method} & CNNs & MobileNets & ResNets \\ 
\midrule
Local Training & 74.83 $\pm$ 1.29 & 65.63 $\pm$ 0.58 & 74.48 $\pm$ 0.46 \\ 
FlexiFed \cite{wang2023flexifed} & 75.03 $\pm$ 1.32 & 64.56 $\pm$ 1.24 & 72.14 $\pm$ 1.58 \\ 
FedGH \cite{yi2023fedgh} & 75.53 $\pm$ 1.12 & 70.32 $\pm$ 1.35 & 76.81 $\pm$ 0.34 \\ 
DPFNAS(no NAS) & \textbf{79.29 $\pm$ 1.02} & \textbf{73.68 $\pm$ 1.08} & \textbf{77.84 $\pm$ 1.06} \\ 
\bottomrule
\end{tabular}
}
\label{tab:accuracy_comparison}
\end{table}

\subsubsection{\textbf{Flexibility of block-based search space and GA-based architecture search mechanism}}
We further test the flexibility of our proposed block-based search space and GA-based architecture search mechanism under different degrees of statistical heterogeneity. Specifically, we simulate both high and low non-IID settings using the CIFAR-100 dataset, controlling the number of label categories and the sampling skew per client.

\begin{figure}[t!]
    \centering
    \subfloat[\textnormal{\fontfamily{ptm}\selectfont\footnotesize High Non-IID}\label{fig:HighNon-iid}]{
        \includegraphics[width=1\linewidth]{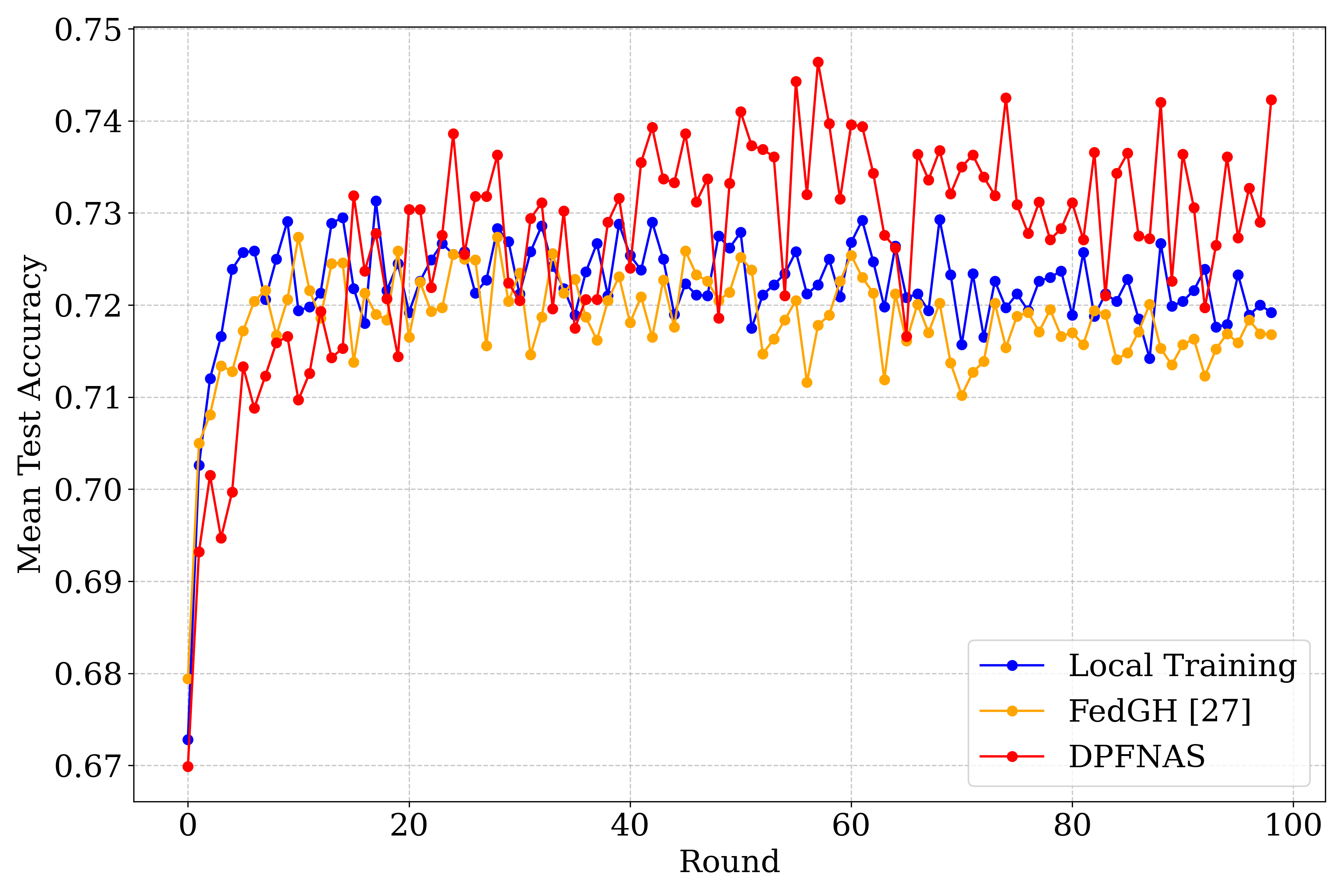}
    }
    \hfill
    \subfloat[\textnormal{\fontfamily{ptm}\selectfont\footnotesize Low Non-IID}\label{fig:LowNon-iid}]{
        \includegraphics[width=1\linewidth]{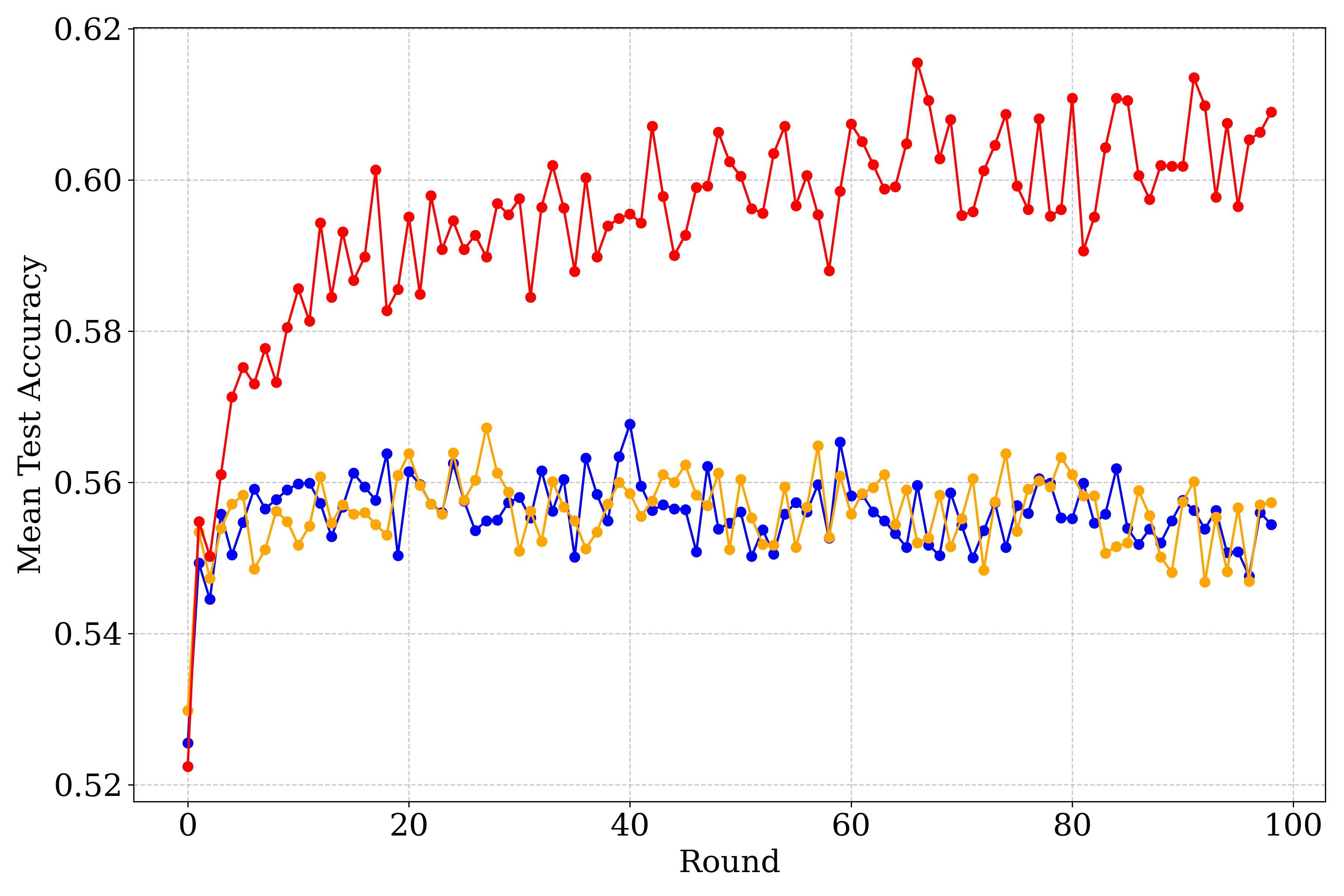}
    }
    \caption{Test accuracy under high and low non-IID data distributions.}
    \label{fig:Non-iid_comparison}
\end{figure}

As shown in Fig.~\ref{fig:Non-iid_comparison}, our proposed PANAS consistently achieves better and more stable convergence than competing baselines in both settings. In the low non-IID case as shown in Fig.~\ref{fig:LowNon-iid}, the proposed PANAS outperforms others by up to 5\% and demonstrates early convergence starting from the third communication round. In the high non-IID case as shown in Fig.~\ref{fig:HighNon-iid}, where client distributions are more disjoint, the proposed PANAS maintains its superiority with about 3\% improvement over FlexiFed \cite{wang2023flexifed} and FedGH \cite{yi2023fedgh}, while also exhibiting smooth and stable training curves without divergence. These results confirm that our proposed DPFNAS is resilient to client data imbalance and skewed distributions, a key advantage in realistic federated deployments.

\subsubsection{\textbf{Effectivness of BO-based hyperparameter optimization strategy}}
To assess the effectiveness of our proposed BO-based hyperparameter optimization strategy in improving model utility under different privacy guarantees, we compare the full algorithm with a variant in which the hyperparameter search component is removed. Specifically, both versions adopt the same privacy-preserving training pipeline, but the baseline variant uses fixed hyperparameter configurations manually specified in prior work. These configurations include the local learning rate ($\eta_w$), noise scale ($\sigma$), clipping tehreshold ($C$), and batch size ($B$). The full hyperparameter settings for the fixed-configuration variant are summarized in Table~\ref{tab:hyperparameter_config}.

The full DP-SGD hyperparameter configurations used in our experiments are summarized in Table~\ref{tab:hyperparameter_config}.
We evaluate five baseline variants (Fixed Config A–E), each combining NAS with manually specified hyperparameter settings commonly used in prior works \cite{jayaraman2019evaluating}.
As a comparison, the proposed DPFNAS optimizes both the architecture and the hyperparameters under a fixed privacy budget of $\epsilon = 5.0$.
The hyperparameter search is conducted on a representative client using a held-out validation set, and runs for 30 iterations to balance search efficiency and computational cost.

\begin{table}[t!]
\caption{ DP-SGD parameter configurations: five fixed hyperparameter baselines vs. BO-based hyperparameter optimization.}
\centering
\resizebox{\columnwidth}{!}{
\begin{tabular}{l>{\raggedleft\arraybackslash}p{1.5cm}>{\raggedleft\arraybackslash}p{1cm}>{\raggedleft\arraybackslash}p{1cm}>{\raggedleft\arraybackslash}p{1cm}}
\toprule
\textbf{Config}     & \textbf{$\eta_w$}   & \textbf{$\sigma$} & \textbf{$C$} & \textbf{$SR$} \\ 
\midrule
Fixed Config A   & 0.0100   & 1.00                 & 10.00              & 0.20          \\ 
Fixed Config B   & 0.0100   & 1.00                 & 1.00               & 0.20          \\ 
Fixed Config C   & 0.0100   & 2.00                 & 10.00              & 0.20          \\ 
Fixed Config D   & 0.0100   & 1.00                 & 10.00              & 0.02          \\ 
Fixed Config E   & 0.0001   & 1.00                 & 10.00              & 0.20          \\ 
DPFNAS & \textbf{0.0010} & \textbf{1.92} & \textbf{0.50} & \textbf{1.00} \\
\bottomrule
\end{tabular}
}
\label{tab:hyperparameter_config}
\end{table}

\begin{figure}[t!]
    \centering
    \includegraphics[width=1\linewidth]{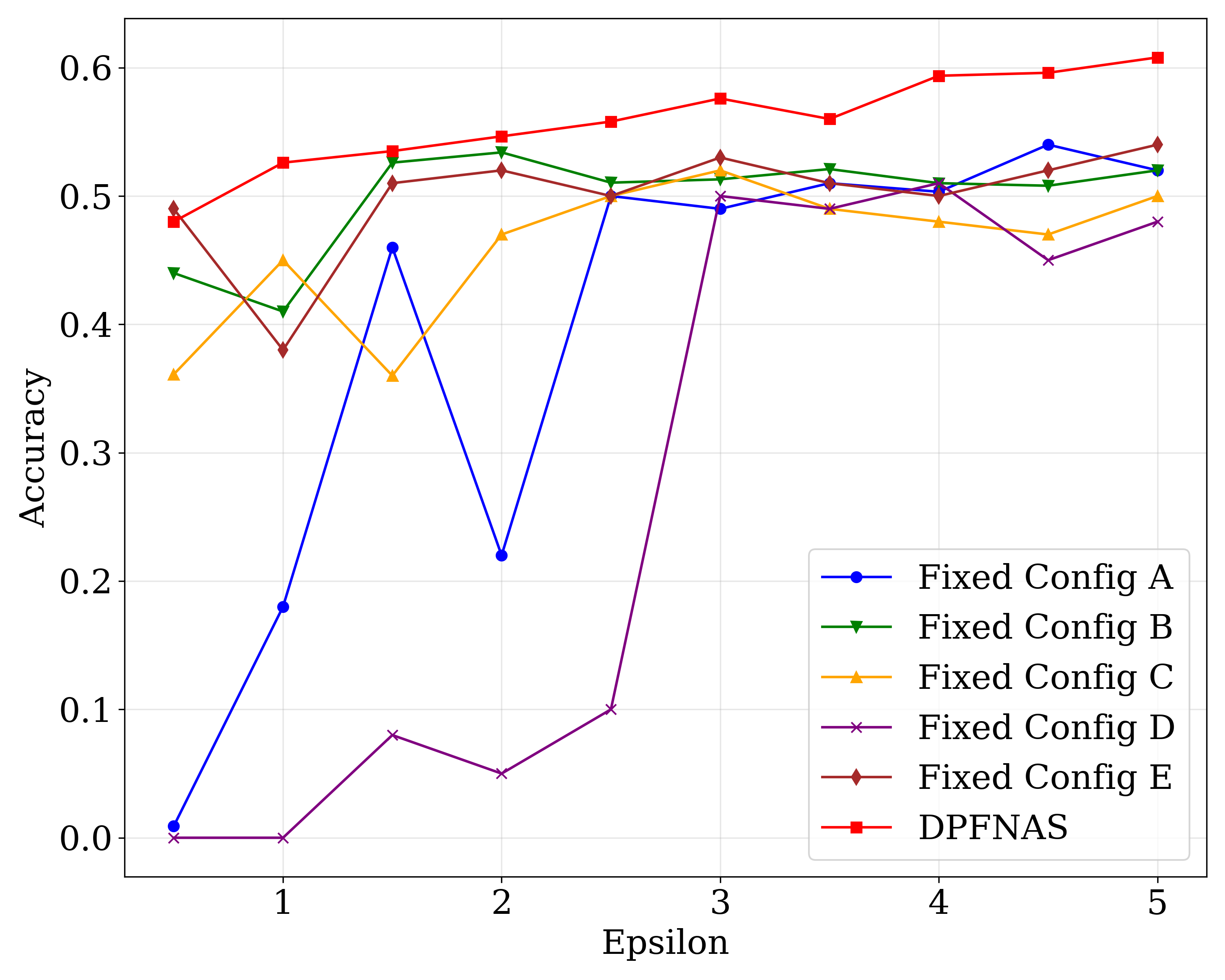}
    \caption{Accuracy under different privacy costs. DPFNAS achieves consistently higher utility than manually tuned baselines across all privacy levels.}
    \label{fig:Utiliy vs Privacy}
\end{figure}

To quantify the effectiveness, we evaluate model performance under varying privacy budgets $\epsilon$. Fig.~\ref{fig:Utiliy vs Privacy} shows the comparison of the test accuracy of privacy-aware hyperparameter optimization and the manual baselines across different privacy levels.

From the Fig.~\ref{fig:Utiliy vs Privacy}, the proposed BO-based hyperparameter optimization strategy consistently yields superior performance across all tested values of $\epsilon$. For instance, when $\epsilon = 1.0$, it improves accuracy by approximately 10\% over the strongest manually tuned baseline. Even under looser privacy constraints ($\epsilon = 5.0$), it still maintains a 6\% accuracy advantage. These improvements highlight the importance of jointly optimizing DP parameters rather than relying on heuristic selection.

The observed improvements in test accuracy under privacy constraints are primarily attributed to the proposed BO-based hyperparameter optimization strategy, which effectively balances key factors such as learning rate, noise scale, and gradient clipping. Guided by both theoretical insights and empirical evidence, this optimization reduces the adverse effects of noise injection while strictly preserving differential privacy guarantees. These results confirm that jointly tuning privacy-related hyperparameters is crucial for achieving superior model utility in DP-SGD-based privacy-preserving training.

\section{Conclusion}

In this paper, we proposed DPFNAS, a novel privacy-preserving and adaptive federated learning framework designed for the heterogeneous and privacy-sensitive environments of 6G edge networks. To enhance architectural adaptability across diverse client data distributions, we develop PANAS, a flexible NAS algorithm that leverages a lightweight block-based search space and combines Genetic Algorithms with Bayesian Optimization for privacy-aware architecture optimization. To address the privacy leakage risks in conventional supernet-based FNAS, we further  introduce DPRBFT, a sample-level representation-based federated training strategy that avoids direct parameter sharing by aggregating intermediate features and incorporates personalized DP-SGD to ensure differential privacy guarantee. We theoretically analyze the convergence of our differentially private training process and conduct extensive experiments on CIFAR-10 and CIFAR-100. The experimental results demonstrate the effectiveness of DPFNAS in balancing accuracy, communication efficiency, and privacy. This work offers a promising direction for future research on secure, efficient, and personalized federated learning frameworks for next-generation intelligent network systems.

\section*{Acknowledgment}

This work was supported in part by the National Natural Science Foundation of China (No. U23B2024), the National Key Research and Development Program of China (No. 2022YFB2902203), and the Guangdong Basic and Applied Basic Research Foundation (No. 2024A1515011492).

\bibliographystyle{ieeetr}
\bibliography{IEEEabrv,references}

\end{document}